\documentclass[10pt,twocolumn,letterpaper]{article}

\usepackage{cvpr}
\usepackage{times}
\usepackage{graphicx}
\usepackage{amsmath}
\usepackage{amssymb}
\usepackage{balance}
\usepackage{multirow}
\usepackage{xspace}
\usepackage{multirow}
\usepackage{verbatim}
\usepackage{enumerate}
\usepackage{enumitem}

\usepackage[usenames,dvipsnames]{color}  
\usepackage[pagebackref=false,breaklinks=true,colorlinks,bookmarks=false,urlcolor=blue,linkcolor=blue,citecolor=blue,pdftitle={Improved Anomaly Detection in Crowded Scenes via Cell-based Analysis of Foreground Speed, Size and Texture},pdfauthor={Vikas Reddy, Conrad Sanderson, Brian C. Lovell}]{hyperref}

\def\Vec#1{{\boldsymbol{#1}}}
\def\Mat#1{{\boldsymbol{#1}}}

\cvprfinalcopy

\def\cvprPaperID{xx}

\begin{document}

\title
  {
  Improved Anomaly Detection in Crowded Scenes\\
  via Cell-based Analysis of Foreground Speed, Size and Texture
  }

\author
  {
  Vikas Reddy,
  Conrad Sanderson,
  Brian C. Lovell
  \\
  ~\\
  NICTA, PO Box 6020, St Lucia, QLD 4067, Australia \thanks{{\bf Published~in:} IEEE Conference on Computer Vision and Pattern Recognition Workshops (CVPRW), pp.~55--61, 2011. \href{http://dx.doi.org/10.1109/CVPRW.2011.5981799}{http://dx.doi.org/10.1109/CVPRW.2011.5981799}}
  \\
  The University of Queensland, School of ITEE, QLD 4072, Australia
  }

\maketitle
\thispagestyle{empty}

\begin{abstract}

\vspace{-2ex}
\noindent
A robust and efficient anomaly detection technique is proposed,
capable of dealing with crowded scenes
where traditional tracking based approaches tend to fail.
Initial foreground segmentation of the input frames confines the analysis to foreground objects
and effectively ignores irrelevant background dynamics.
Input frames are split into non-overlapping cells,
followed by extracting features based on motion, size and texture from each cell.
Each feature type is independently analysed for the presence of an anomaly.
Unlike most methods, a refined estimate of object motion is achieved by computing the optical flow of only the foreground pixels.
The motion and size features are modelled by an approximated version of kernel density estimation,
which is computationally efficient even for large training datasets.
Texture features are modelled by an adaptively grown codebook,
with the number of entries in the codebook selected in an online fashion.
Experiments on the recently published UCSD Anomaly Detection dataset 
show that the proposed method obtains considerably better results 
than three recent \mbox{approaches}:
\mbox{MPPCA}, \mbox{social~force}, and mixture of dynamic textures~(MDT).
The~proposed method is also several orders of magnitude faster than~MDT,
the next best performing method.

\end{abstract}

\vspace{-1ex}
\section{Introduction}
\vspace{-0.5ex}

Automated detection of anomalous events in video feeds has the potential
to provide more vigilant surveillance,
possibly in lieu of,
or as an assistance to,
human operators
who have limited attention spans when faced with tedious tasks~\cite{haering2008evolution}.
Qualifying an event as anomalous is subjective and depends on the intended application as well as context.
However, without being application or context specific,
an anomalous event can be defined as
{\it any event that is different from what has been observed beforehand}.

Detection of anomalous events can be hence viewed as a binary classification problem,
where there are training examples only for one class, generally.
Typical algorithms model the dynamics of `normal' activity or `expected' behaviour
and compare new observations with an existing model. 
Any outliers are labelled as anomalous. 
An ideal system is expected not only to detect anomalous events accurately,
but also to adapt itself to the changes witnessed in the environment over time.

Several anomaly detection techniques have been proposed in various research fields.
Chandola et al.~\cite{chandola2009anomaly} discuss them in detail in their survey,
while Saligrama et al.~\cite{saligrama2010video}
examine video based anomaly detections approaches in the context of surveillance.
Existing methods in the literature can be roughly placed into two categories: 
{\bf (i)} analysis by tracking,
where trajectories of individual objects are maintained;
{\bf (ii)} analysis without tracking,
where other features such as motion and texture are employed to model activity patterns of a given scene.

In the first category,
almost all approaches use tracking information
directly to gather object speed and direction,
and indirectly as an aid in determining features
such as the size and aspect ratio of objects~\cite{basharat2008learning,hu2006system,piciarelli2008trajectory,remagnino2001classifying,wang2006learning}.
While trajectory based approaches are suitable for cases where the scene is comprised of only a few objects,
in crowded environments it is difficult to reliably maintain tracks due to occlusion and overlap of objects~\cite{kratz2009anomaly,mahadevan2010anomaly}.


In light of the above problems,
in the second category
the anomaly detection task is formulated while deliberately omitting the tracking of specific objects.
Most approaches in this category largely rely on motion or motion-related features. 
For example, 
Mehran et al.~\cite{mehran2009abnormal} model crowd behaviour using
a ``social force" model,
where the interaction forces are computed using optical flow.
Adam et al.~\cite{adam2008robust} model optical flow at a set of fixed spatial locations using probabilistic histograms.
Ermis et al.~\cite{ermis2008motion}~propose using busy-idle rates of each pixel to detect abnormal behaviour.


As the above techniques solely rely on motion information,
anomalies occurring due to object size or appearance may not be detected. 
To address this limitation,
Mahadevan et al.~\cite{mahadevan2010anomaly} recently proposed to jointly model the appearance and dynamics of crowded scenes,
using mixtures of dynamic textures (MDT)~\cite{Chan_MDT_2008}.
The method explicitly investigates both temporal and spatial anomalies.
Though the reported comparative results show improvements over earlier techniques,
the method's main drawback is heavy computation.
Evaluating a frame of size {\small $240 \times 160$} takes about {\small $25$} seconds (ie.~{\small $2.4$} frames per minute).


In this paper, we present a robust anomaly detection algorithm with relatively low complexity, 
targeted primarily for crowded scenes where traditional tracking based approaches tend to fail.
To suppress undesirable background dynamics,
such as waving trees and illumination variations, 
we perform foreground segmentation and retain only foreground objects for further analysis.
Each input frame is split into non-overlapping cells (small regions).
Based on each frame's foreground mask,
the relevant cells are analysed for the presence of an anomaly.
Unlike most methods, a refined estimate of motion is achieved by computing the optical flow only for the foreground pixels.
In addition to motion,
the proposed method analyses the size and texture of foreground objects at each cell location.


Motion, size and texture are modelled separately.
Independent analysis helps to keep computation efficient,
and allows for inferring the nature of the anomaly
(eg.~speed violation, lack of motion, size too large, etc).
Each cell is labelled as either normal or anomalous
after combining the outputs of multiple classifiers (one for each feature type).


We continue the paper as follows.
In Section~\ref{sec:Proposed Algorithm} the proposed algorithm is described in detail.
Performance evaluation and comparison with three recent algorithms is given in Section~\ref{sec:Experiment Results}.
The main findings and possible future directions are presented in Section~\ref{sec:conclusion}.

\section {Proposed Algorithm}
\label{sec:Proposed Algorithm}


\noindent
The proposed method has four main components:

\setenumerate[0]{leftmargin=2.8ex,itemindent=0ex}

\begin{enumerate}
\setlength{\itemsep}{0ex}

\item
Feature extraction,
where input images are split into non-overlapping spatial regions,
termed {\it cells},
and features are extracted based on motion, size and texture
of the foreground objects contained in the cells.

\item
Model estimation, which models the normal dynamics witnessed at each cell location.
There are separate models for each feature type.

\item
Classification of each cell as anomalous or normal,
where each cell is sequentially checked for normality by up to two classifiers.
As soon as the first classifier deems that the cell is anomalous,
the second classifier is~not consulted. 
In order of processing, the two classifiers are:

\begin{small}
\setenumerate[0]{leftmargin=3.8ex,itemindent=0ex}
\begin{enumerate}

\item
Speed check,
where the likelihood of the magnitude of motion of any foreground objects is evaluated.

\item
Size and texture check,
where 
first the likelihood of the size of a foreground object is evaluated.
Cells with low likelihoods (suggesting an anomaly)
are further analysed according to their texture,
in order to validate the presence of the anomaly.  

\end{enumerate}
\end{small}

\item
Spatio-temporal post-processing,
to minimise isolated random noise present in the generated anomaly masks.

\end{enumerate}

\noindent
Each of the components is explained in detail in the following sections.
\subsection {Feature Extraction}
\label{subsec:feature_extraction}
\vspace{-0.5ex}

Let the resolution of the greyscale image sequence {\small $\mathcal{I}$} be {\small $\mathcal{W} \times \mathcal{H}$}.
Each image is split into non-overlapping cells (\mbox{regions}) of size {\small $N \times N$},
with the cell located at {\small $i$} and {\small $j$} denoted by {\small $\Mat{C}{(i,j)}$}.
The cell co-ordinates have the range of
\mbox{\small $i = 1,2,\cdots,(\mathcal{W}/N)$}
and
\mbox{\small $j = 1,2,\cdots,(\mathcal{H}/N)$}.
Let {\small $\Mat{I}_t$} be the frame at time instant {\small $t$} 
and let its corresponding cells be denoted by {\small $\Mat{C}_t(i,j)$}.

In order to restrict the analysis to regions of interest and to filter out distractions
(eg.~waving trees, illumination changes, etc),
we perform foreground segmentation on each incoming frame.
We have used the method proposed in~\cite{reddy2010adaptive},
due to its robustness, high-quality foreground masks
and the ability to estimate the background even in the presence of multiple moving foreground objects.
Alternative techniques for estimating the background in crowded scenes include~\cite{Baltieri_2010,Reddy_IVP_2011}.

For each cell, we extract features based on motion, size and texture.
The foreground masks are referenced while computing the features. 
The details of the three features are given below.

\subsubsection{Motion}
\label{subsubsec:motion}
\vspace{-0.5ex}

To estimate the motion associated with cell {\small $\Mat{C}_t{(i,j)}$}, 
we compute the optical flow of only the foreground pixels.
The iterative Lucas-Kanade algorithm~\cite{bouguet1999pyramidal,lucas1981iterative}
is employed to compute the displacement of pixels between two consecutive frames,
with a fixed search window around each pixel.
We first calculate the average motion associated with cell {\small $\Mat{C}_t{(i,j)}$} using:

\vspace{-1ex}
\begin{small}
\begin{equation}
  \widehat{\operatorname{mot}}_t(i, j)
  =
  \frac{1}{N_f} \sum\nolimits_{n=0}^{N_f} \left|\hspace{-0.4ex}\left|  \left[ v_{x}^{(n)}, ~ v_{y}^{(n)} \right]  \right|\hspace{-0.4ex}\right|_1
  \label{eqn:motion}
\end{equation}%
\end{small}%

\noindent
where, for foreground pixel {\small $n$}, {\small $v_{x}^{(n)}$} and {\small $v_{y}^{(n)}$}
are the optical flows in the {\small $x$} and {\small $y$} directions, respectively,
while {\small $N_f$} is the total number of foreground pixels within the cell.


The motion feature for cell {\small $\Mat{C}_t{(i,j)}$} is taken to be the smoothed (noise-reduced) version of the cell's average motion,
calculated using straightforward temporal averaging:

\vspace{-1ex}
\begin{small}
\begin{equation}
  {\operatorname{mot}}_t(i, j) =  \frac{1}{3} \sum\nolimits_{u = t-1}^{t+1} \widehat{\operatorname{mot}}_{u}(i, j)
  \label{eqn:tmprl_motion}
\end{equation}%
\end{small}%

\subsubsection{Size} 
\label{subsubsec:size_and_texture} 
\vspace{-0.5ex}

Relying on motion alone for anomaly detection might be insufficient,
as certain anomalies may exhibit motion that is considered as normal
(eg.~a slow moving vehicle on a path designated for pedestrians). 
Furthermore, motion estimation techniques can suffer from the aperture problem~\cite{trucco1998introductory}.
To increase the sensitivity of anomaly detection, the size of foreground objects can be analysed.

\begin{figure}[!b]
  \begin{minipage}{1\columnwidth}
    \includegraphics[width=0.32\columnwidth]{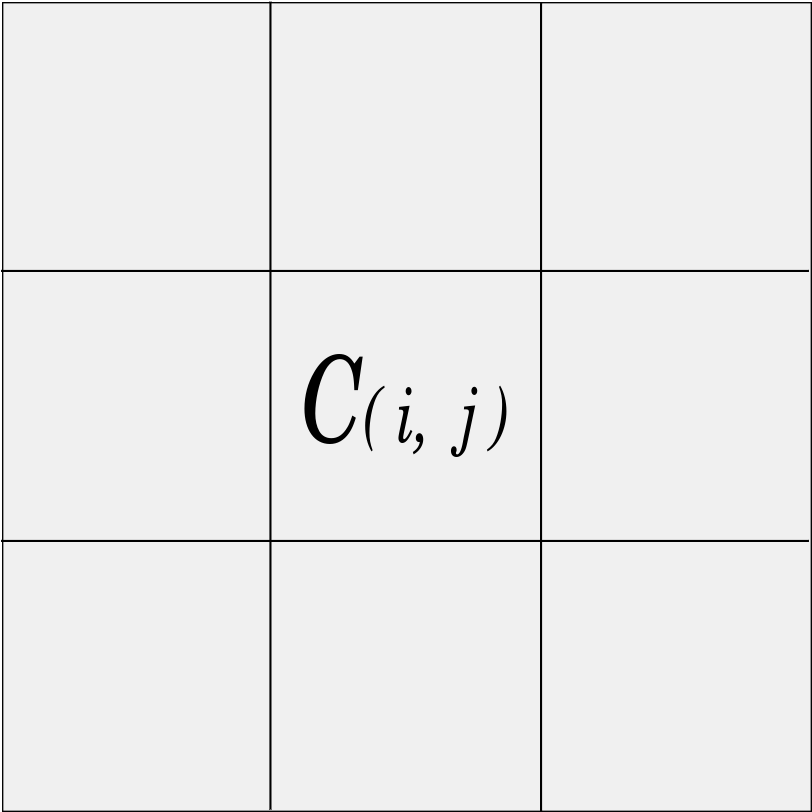}
    \hfill
    \includegraphics[width=0.32\columnwidth]{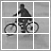}
    \hfill
    \includegraphics[width=0.32\columnwidth]{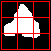}
  \end{minipage}
 
 \vspace{0.5ex}

  \begin{minipage}{1\columnwidth}
    \begin{minipage}{0.32\columnwidth}
      \centerline{\small\bf (a)}
    \end{minipage}
    \hfill
    \begin{minipage}{0.32\columnwidth}
      \centerline{\small\bf (b)}
    \end{minipage}
    \hfill
    \begin{minipage}{0.32\columnwidth}
      \centerline{\small\bf (c)}
    \end{minipage}
  \end{minipage}

 
  \caption
    {
    \small
    {\bf(a)}~the approximate size of the foreground object contained in the center cell {\footnotesize $\Mat{C}{(i,j)}$}
    is computed by considering foreground pixels in the cell as well as its neighbouring cells;
    {\bf(b)}~an example of a foreground object appearing in the center cell and its neighbours;
    {\bf(c)}~the corresponding foreground mask,
    found via an automatic foreground segmentation algorithm.
    }
  \label{fig:neighbourhood}
\end{figure}

A common technique to measure object size is via connected component analysis on the foreground masks.
However, in crowded environments it becomes ineffective due to object overlap and occlusion.
Instead, an approximate size of an object contained within a cell
can be obtained by considering its foreground occupation 
along with that of its neighbouring cells (as the object may occupy more than one cell).
An example is shown in Fig.~\ref{fig:neighbourhood}.



Specifically,
let us denote the foreground occupancy (number of foreground pixels) for cell {\small $\Mat{C}_t{(i,j)}$} by {\small ${o}_t{(i,j)}$}.
We define the size feature for cell {\small $\Mat{C}_t{(i,j)}$}
as a weighted combination of the foreground occupancy values of the cell and its immediate neighbours:

\vspace{-2ex}
\begin{small}
\begin{equation}
  \operatorname{size}_t{(i,j)}
  = 
  \sum_{a = i\mbox{-}1}^{i+1} ~ \sum_{b = j\mbox{-}1}^{j+1}
  G(a\mbox{~-~}i\mbox{~+~}1, ~ b\mbox{~-~}j\mbox{~+~}1) ~ {o}_t{(a,b)} 
  \label{eqn:size}
\end{equation}%
\end{small}%

\noindent
where {\small $G$} is a {\small $3 \times 3$} Gaussian mask~\cite{Gonzalez_2008}.
The mask is used for placing prominence on the center cell
and hence reducing the impact of neighbouring cells that,
in crowded scenarios,
may contain foreground pixels belonging to other objects (in addition to the object of interest).




\subsubsection{Texture}


While the size feature can be useful for increasing the sensitivity of anomaly detection,
using it without qualification may also increase the false alarm rate.
For example,
in crowded environments the foreground masks of people walking close to each other could resemble a large foreground object.

To address this problem,
the texture present within the cell can be used for increasing selectivity.
To this end, we filter a given image using 2D Gabor wavelets~\cite{lee2002image}
at four orientations: {\small $0$}, {\small $45$}, {\small $90$} and {\small $135$} degrees.
The texture descriptor for cell {\small $\Mat{C}_t{(i,j)}$} is hence a 4D vector:

\vspace{-1ex}
\begin{small}
\begin{equation}
  \operatorname{\bf txt}_t{(i,j)} =  \left[ m_0 ~~ m_{45} ~~ m_{90} ~~ m_{135} \right]'
  \label{eqn:texture}
\end{equation}%
\end{small}%

\noindent
where
{\small $m_\theta$} is the sum of the response magnitudes of the wavelet oriented at {\small $\theta$} degrees,
over the pixels contained within the cell.
The texture vectors are only collected for cells that have at least one foreground pixel,
in order to minimise modelling of the background.
\subsection{Scalable Semi-Parametric Model Estimation}
\label{subsec: Model Estimation}

Surveillance scenarios include platforms at train/bus stations,
buildings (both indoors/outdoors) as well as road/walkway traffic.
In all these scenarios,
even normal day-to-day events have inherent variations that are random in nature. 
For example, the speed of vehicles on a road can vary arbitrarily due to traffic light signals and congestion.
Furthermore, the dynamics of the scene can keep changing over time.
As such, parametric approaches are unlikely to be effective for modelling distributions of features in these scenarios,
as the number of modes is unlikely to be reliably known beforehand.
With this in mind,
we employed a semi-parametric modelling approach,
which can be used for modelling arbitrary distributions 
without using any assumptions on the forms of the underlying densities.

We model each cell by considering only the features extracted from that particular cell location,
along the temporal axis.
As described in Section~\ref{subsec:feature_extraction},
there are three feature types: motion (a scalar), size (also a scalar) and a texture descriptor (a 4D vector).
In tasks such as object detection/recognition,
it is often argued that joint modelling of features yields better results
than modelling them independently~\cite{bahlmann2005system,shotton2006textonboost}.
However, in our context such an approach may fail to detect outliers (ie.~the anomalies) accurately,
due to the implicit mutual influence exhibited by the features in the decision process. 
Furthermore, the dynamics of a crowded scene are inherently arbitrary in nature,
which may render joint modelling ineffective.
We will hence model these features separately.
Independent analysis keeps computation efficient,
examines each feature precisely for anomalies,
which in turn allows for inferring the nature of the anomaly 
(eg.~speed violation, lack of motion, size too large, etc).
 

For modelling the motion and size features,
a straightforward and efficient semi-parametric approach involves constructing normalised histograms.
The training data can be discarded once the histograms are built.
However,
a major problem with this approach is the presence of sharp discontinuities in the estimated densities due to binning,
rather than that of the underlying distribution that generated the data~\cite{bishop2006pattern}.
To overcome the above limitation,
it is possible to use kernel density estimation techniques
that result in smoother probability density functions~\cite{bishop2006pattern,saleemi2008probabilistic}. 
Their training phase only involves storing of all the data samples. 
However, their drawback is increased computational cost and memory requirements
as the dataset becomes larger~\cite{bishop2006pattern}.
As such, these techniques can suffer from scalability issues.

To achieve a better trade-off between accuracy and computational requirements,
we create a smoothed histogram by
temporarily storing all the training samples
and performing Gaussian kernel based density estimation
to compute the probability of the continuous variable (ie.~motion or size)
only at discrete points over its entire permissible range. 
In~effect, we assume the random variable to be discrete
and compute its probability at fixed points using:

\vspace{-2ex}
\begin{small}
\begin{equation}
  f(s\Delta{x})
  =
  \frac{1}{N} 
  \sum\nolimits_{n = 1}^{N} \frac{1}{h \sqrt{2\pi}} \exp \left\{ - \frac{{\parallel s\Delta {x} - {x_n}\parallel}^2}{2h^2} \right\}
  \label{eqn:kd}
\end{equation}%
\end{small}%

\noindent
where {\small $\Delta{x}$} is the resolution of the step size {\small{(eg.~0.25)}},
\mbox{\small $s = \{ 0, 1, 2, 3, ..., S\}$},
with {\small $S\Delta{x}$} being the valid upper limit of the variable in consideration.
{\small $N$} is the number of samples in the training dataset and {\small $h$} is the bandwidth of the Gaussian kernel.
The probability values are normalised to obtain a probability mass function (pmf).
As in the histogram approach,
the training data is discarded once the pmf is computed.
The resultant pmf is denoted by {\small $\widehat{p}(\cdot)$}.


%
%



\subsubsection{Adaptive Modelling of Texture Descriptors}

While the above approach is effective for modelling the distribution of scalars (motion and size in our case),
using it to model the distribution of the 4D texture descriptors is impractical,
as the number of resulting discrete samples required to cover the entire feature space
(for just one cell location) would be quite large.
For example,
having only {\small $20$} equally spaced points in each dimension would generate {\small $20^4$} points in 4D space.
As there is a non-trivial number of cell locations,
the total storage costs would be hence prohibitive.

Furthermore,
the above density estimation approach implicitly relies on an Euclidean based distance,
which will be affected by variations in texture contrasts,
rather than purely measuring the differences between texture patterns.
For example, the texture descriptor will exhibit low magnitude responses
when the intensity of a pedestrian's clothing is similar to that of the background,
and high magnitude responses when the intensity is contrasting to the background.

To address the storage problem,
for each cell location we model the distribution of the texture descriptors using a codebook that is
trained in an online fashion (adaptively grown),
inspired by~\cite{kratz2009anomaly}. 
To address the distance measure problem,
we employ Pearson's correlation coefficient~\cite{theodoridispattern}
for measuring the similarity of two descriptors:

\vspace{-1ex}
\begin{small}
\begin{equation}
  %
  \rho( \Vec{a}, \Vec{b} )
  =
  \frac
  { ( \Vec{a} - \mu_\Vec{a} )' ( \Vec{b} - \mu_\Vec{b} ) }
  { \parallel \hspace{-0.5ex} \Vec{a} - \mu_\Vec{a} \hspace{-0.5ex}  \parallel  \parallel \hspace{-0.5ex} \Vec{b} - \mu_\Vec{b} \hspace{-0.5ex} \parallel }
  \label{eqn:correlation}
\end{equation}%
\end{small}%

\noindent
where {\small $\mu_\Vec{x}$} is the mean of the elements of vector {\small $\Vec{x}$},
and {\small $\rho( \Vec{a}, \Vec{b} ) \in [-1,+1]$}.

The codebook is built as follows.
Initially, the first training vector is taken to be the first entry in the codebook.
Each of the remaining vectors is sequentially treated as a new observation, {\small $\Vec{x}$},
which is compared to each entry in the codebook, {\small $\Vec{c}_k$},
using Eqn.~(\ref{eqn:correlation}).
If, for the best matching {\small $\Vec{c}_{k}$}, {\small $\rho( \Vec{x}, \Vec{c}_{k} ) > 0.9$},
the {\small $k$}-th entry is updated using~\cite{Chan_1979}:

\vspace{-1ex}
\begin{small}
\begin{equation}
  \Vec{c}_{k}^{\operatorname{new}}
  =
  \Vec{c}_{k}^{\operatorname{old}}
  +
   \frac{1}{W_k+1} \left( \Vec{x} - \Vec{c}_{k}^{\operatorname{old}} \right)
  \label{eqn:ad_recursive_mean}
\end{equation}%
\end{small}%

\noindent
where {\small $W_k$} is the number of texture vectors associated so far with entry {\small $k$}.
If  {\small $\forall k~ \rho( \Vec{x}, \Vec{c}_{k} ) \leq 0.9$},
vector {\small $\Vec{x}$} is appended to the codebook,
thereby expanding it.


\subsection{Cell Classification}
\label{subsec: Classification}
\vspace{-0.5ex}

Each cell is sequentially checked whether it is anomalous by up to two classifiers.
As soon as the first classifier deems that the cell is anomalous,
the second classifier is not consulted.
Specifically,
given a decision threshold {\small $T$},
cell {\small $\Mat{C}_t{(i,j)}$} is classified as anomalous
if either of the following two conditions are satisfied:

\setenumerate[0]{leftmargin=4ex,itemindent=0ex}

\vspace{-1ex}
\begin{enumerate}[label={\small\bf({\alph*})}]
\setlength{\itemsep}{-0.5ex}

\item
{\small $\widehat{p}_{\operatorname{mot}}  \left( \operatorname{mot}_t(i,j) \right) < T$}
\label{item:condition1}

\item
{\small $\widehat{p}_{\operatorname{size}} \left( \operatorname{size}_t(i,j) \right) < T$}
~and~
{\small $\rho_{\operatorname{max}} \left( \operatorname{\bf txt}_t(i,j) \right) < 0.9$}
\label{item:condition2}

\end{enumerate}
\vspace{-1ex}

\noindent
where
{\small $\widehat{p}_{\operatorname{mot}} \hspace{-0.6ex} \left(\cdot\right)$}
and
{\small $\widehat{p}_{\operatorname{size}} \hspace{-0.6ex} \left(\cdot\right)$}
are the pmfs calculated in Section~\ref{subsec: Model Estimation},
while
{\small $\rho_{\operatorname{max}} \left( \operatorname{\bf txt}_t(i,j) \right) =$}
{\footnotesize $\max\limits_{k}$}
{\small $\rho \left( \operatorname{\bf txt}_t(i,j), \textbf{c}_{k} \right)$},
ie.~the correlation coefficient of the closest matching codebook entry.


Condition~{\bf (a)} is effectively a speed check,
where speeds that are either slower or faster than `normal' are detected
(note that {\small $\widehat{p}_{\operatorname{mot}}(\cdot)$} can define several `normal' speeds).
In condition~{\bf (b)} both the size and texture are checked.
As the size feature alone is not be able to distinguish between a large object and a collection of small objects (eg.~a crowd of people),
the texture feature is in effect used to verify the presence of an anomaly indicated by the size feature.

The texture feature is only calculated for cells that contain foreground pixels,
in order to avoid modelling the background (which might be dynamic).
As such, the texture feature is suitable for distinguishing among foreground objects.
However, as the feature may end up capturing irrelevant textures
when the cell contains only thin edges of the foreground,
it is used in combination with the size feature
rather than being used alone.



  
\subsection{Spatio-Temporal Post-Processing}
\label{subsec: post_processing}
\vspace{-0.5ex}

\begin{figure}[!b]
\vspace{-3.5ex}
  \centering
  \includegraphics[width=0.525\columnwidth]{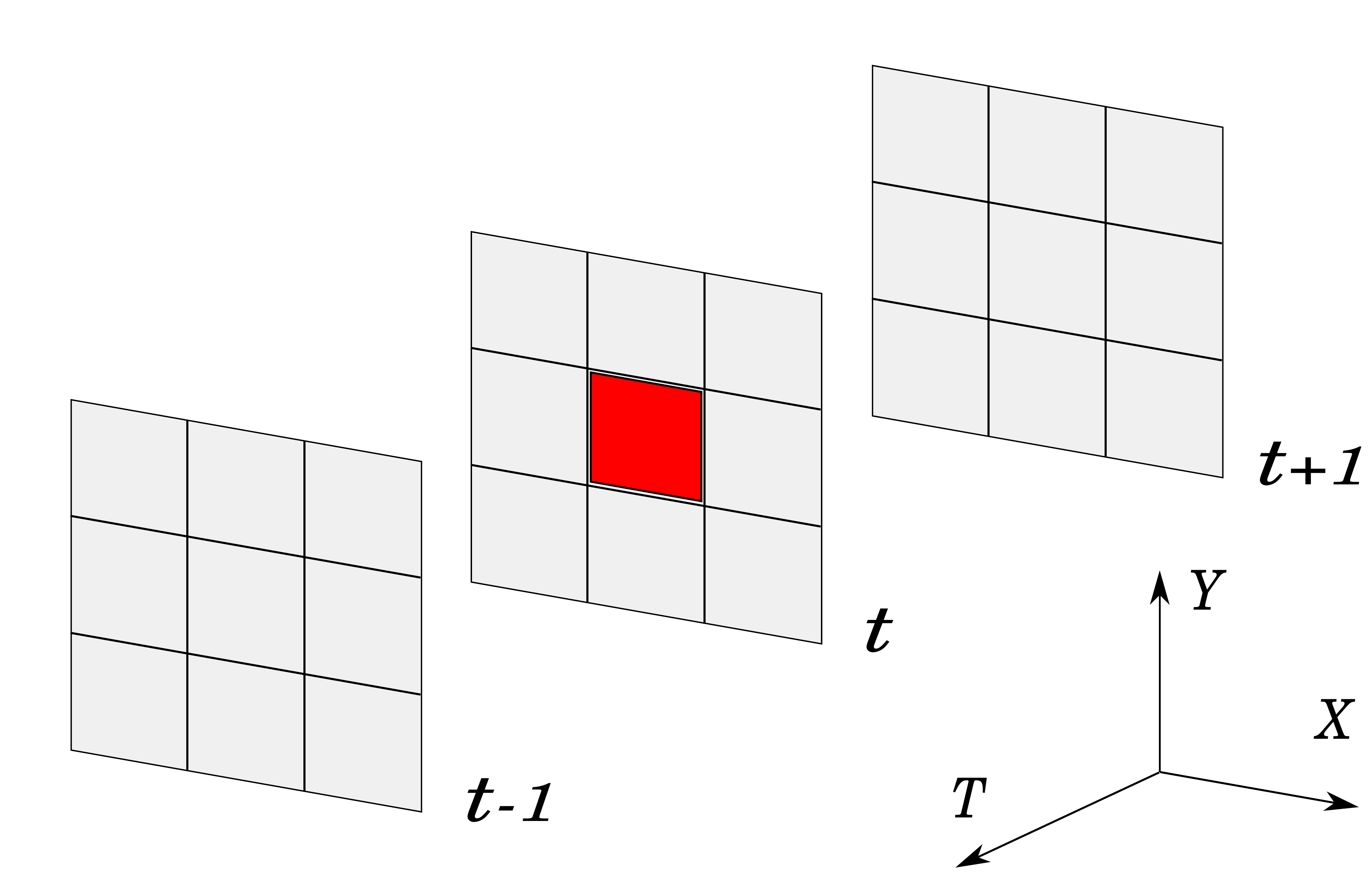}
  \vspace{0.5ex}
  \caption
    {
    For a cell initially classified as anomalous (marked in red),
    its immediate neighbours along both the spatial and temporal axes
    are consulted to verify whether the cell was classified as anomalous due to noise.
    }
  \label{fig:cubepost_proc}
\end{figure}

To minimise spurious and intermittent false alarms,
spatio-temporal post-processing is performed on the anomaly masks
generated by the classification procedure in Section~\ref{subsec: Classification}.
If a cell at time {\small $t$} was initially classified as anomalous,
we consider its immediate neighbours along both the spatial and temporal axes (see Fig.~\ref{fig:cubepost_proc}).
If at least two cells in each plane (ie.~{\small $t\mbox{-}1$}, {\small $t$}, {\small $t\mbox{+}1$}) were classified as anomalous,
we assume that the cell in question was correctly classified as anomalous.
Otherwise, it is re-classified as being normal (ie.~non-anomalous).

\section{Experiments}
\label{sec:Experiment Results}
\vspace{-0.75ex}

\begin{figure}[!t]
\centering
  \begin{minipage}{1\columnwidth}
  \begin{minipage}{1\columnwidth}
    \includegraphics[width=0.49\columnwidth,height = 0.32\columnwidth]{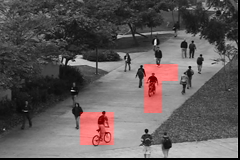}
    \hfill
    \includegraphics[width=0.49\columnwidth,height = 0.32\columnwidth]{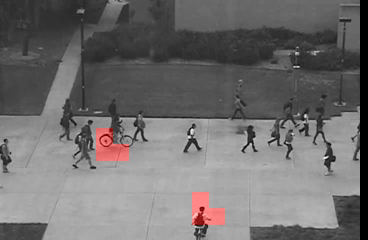}
  \end{minipage}

  \vspace{0.5ex}

  \begin{minipage}{1\columnwidth}
    \includegraphics[width=0.49\columnwidth,height = 0.32\columnwidth]{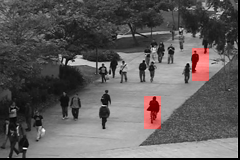}
    \hfill
    \includegraphics[width=0.49\columnwidth,height = 0.32\columnwidth]{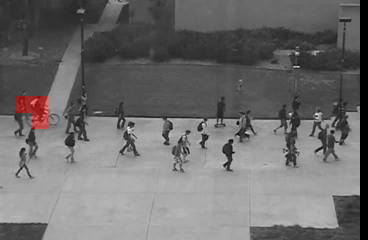}
  \end{minipage}

  \vspace{0.5ex}

  \begin{minipage}{1\columnwidth}
    \includegraphics[width=0.49\columnwidth,height = 0.32\columnwidth]{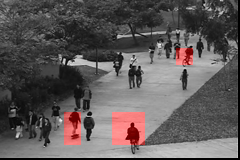}
    \hfill
    \includegraphics[width=0.49\columnwidth,height = 0.32\columnwidth]{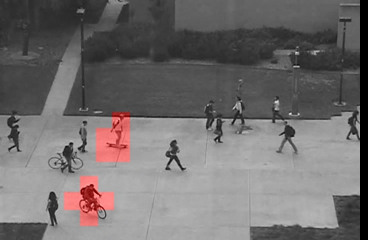}
  \end{minipage}

  \vspace{0.5ex}

  \begin{minipage}{1\columnwidth}
    \includegraphics[width=0.49\columnwidth,height = 0.32\columnwidth]{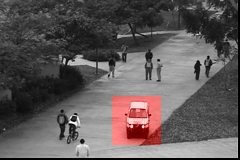}
    \hfill
    \includegraphics[width=0.49\columnwidth,height = 0.32\columnwidth]{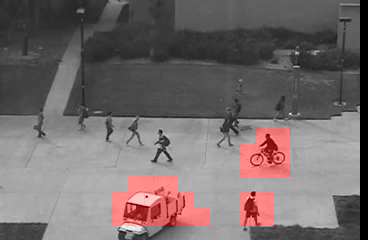}
  \end{minipage}

  \end{minipage} 
  
  \vspace{0.5ex}

\begin{minipage}{1\columnwidth}
   \begin{minipage}{0.49\columnwidth}
   \centerline{\bf(a)}
   \end{minipage}
   \hfill
   \begin{minipage}{0.49\columnwidth}
   \centerline{\bf(b)}
   \end{minipage}
\end{minipage}

  \caption
    {
    Examples of anomaly detection and localisation via the proposed method (highlighted in red).
    Results are shown on the {\bf (a)}~Ped1 and {\bf (b)}~Ped2 subsets of the UCSD Anomaly dataset.
    }
  \label{fig:ad_sample_frames}
  \vspace{-3ex}
\end{figure}

To appraise the performance of the proposed approach,
we performed experiments on the recently released UCSD Anomaly Detection dataset~\cite{mahadevan2010anomaly}.
The dataset contains multiple surveillance videos of two scenes (Ped1 and Ped2),
both with considerable crowds.
Anomalies present in the dataset include:
skateboarders, bikers, motor vehicles, people pushing carts as well as walking on the lawn.
The image size in Ped1 is {\small $238\times158$} pixels,
while on Ped2 it is {\small $360\times240$}.
Ped1 has 34 training and 36 test image sequences,
while Ped2 has 16 training and 12 test image sequences.
Examples are shown in Fig.~\ref{fig:ad_sample_frames}. 

The UCSD dataset has a prescribed evaluation protocol~\cite{mahadevan2010anomaly},
involving two types of evaluations:
{\bf (i)} frame-level anomaly detection,
and
{\bf (ii)} within-frame anomaly localisation.
For frame-level anomaly detection,
all test sequences have annotated groundtruth at frame-level
in the form of a binary flag indicating the presence or absence of anomaly in each frame.
%
For within-frame anomaly localisation,
a subset of test sequences (10 in Ped1 and 9 in Ped2)
has the anomalous regions within each frame marked.
If at least 40\% of detected pixels (belonging to a detected anomaly) match the ground-truth pixels,
it is presumed the anomaly has been localised correctly;
otherwise it is treated as a `miss'. 

The proposed algorithm was compared with methods based on
social force~\cite{mehran2009abnormal},
MPPCA~\cite{10.1109/CVPRW.2009.5206569}, 
and mixture of dynamic textures (MDT)~\cite{mahadevan2010anomaly}.
The first two methods rely on features obtained from optical flow 
while the last approach employs features based on appearance and scene dynamics.
The quantitative results of the above three algorithms were adapted from~\cite{mahadevan2010anomaly}.
To aid the interpretation of the results,
we have reported the false negative rate~\cite{bengio_icml_2005}
instead of the true positive rate used in~\cite{mahadevan2010anomaly}.

Based on preliminary experiments,
the cell size was set to {\small $16\times16$},
while the search window size in the optical flow computation (Sec.~\ref{subsubsec:motion}) was set to {\small $15\times15$}
(odd-sized to ensure a symmetrical search area around a given pixel).
The experiments were implemented with the aid of the \mbox{Armadillo} C++ library~\cite{Armadillo_2010}.

\begin{figure}[!tb]
  \begin{minipage}{1\columnwidth}
    \centering
    
    \begin{minipage}{\textwidth}
      \centering
      \begin{minipage}{0.05\textwidth}
        \centerline{\small\bf (a)}
      \end{minipage}
      \hfill
      \begin{minipage}{0.90\textwidth}
        \centerline{\includegraphics[width=1\textwidth]{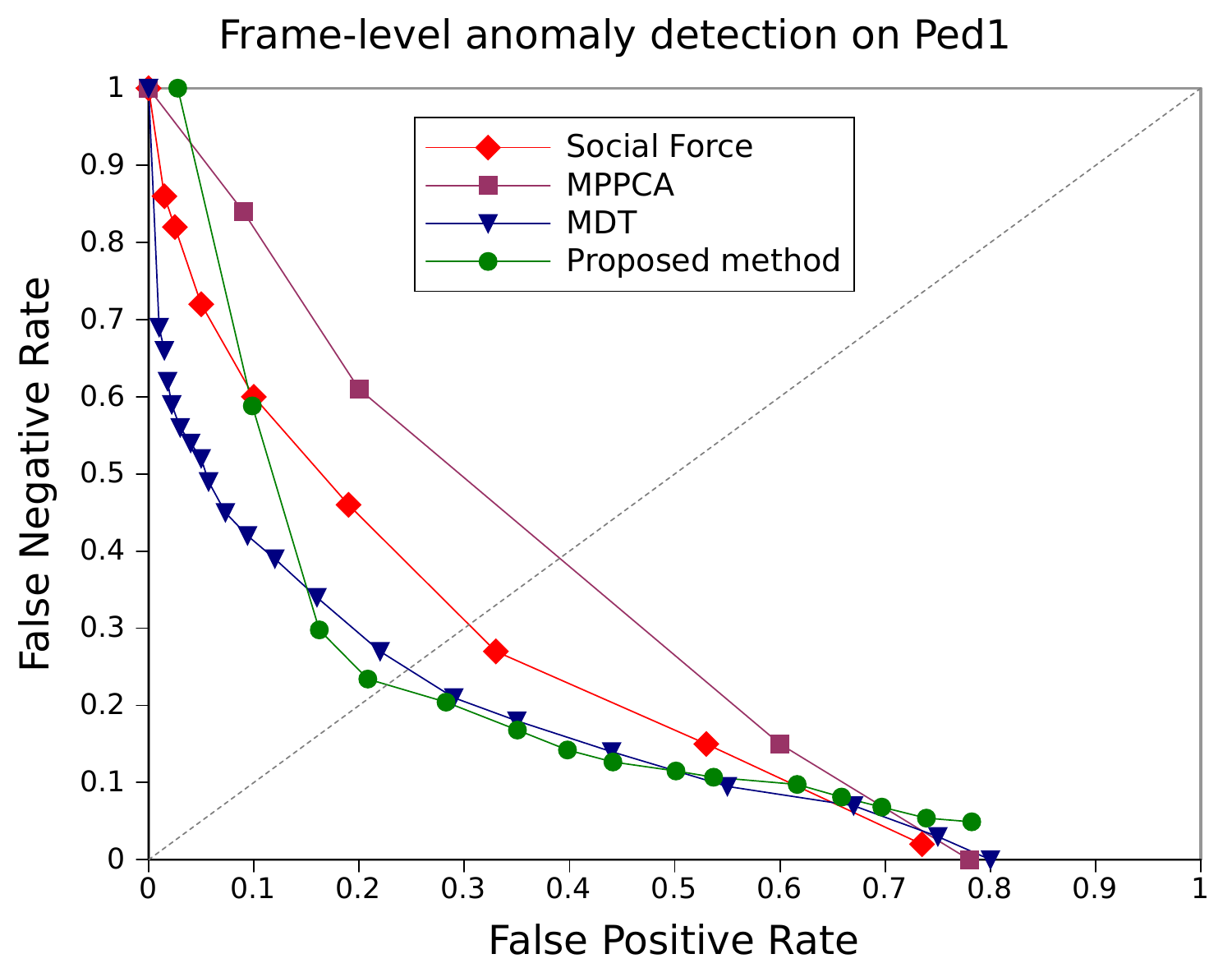}}
      \end{minipage}
    \end{minipage}
    
    \begin{minipage}{\textwidth}
      \centering
      \begin{minipage}{0.05\textwidth}
        ~
      \end{minipage}
      \hfill
      \begin{minipage}{0.90\textwidth}
      ~
      
      ~
      \hrule
      ~
      
      ~
      \end{minipage}
    \end{minipage}
  
    \begin{minipage}{\textwidth}
      \centering
      \begin{minipage}{0.05\textwidth}
        \centerline{\small\bf (b)}
      \end{minipage}
      \hfill
      \begin{minipage}{0.90\textwidth}
        \centerline{\includegraphics[width=1\textwidth]{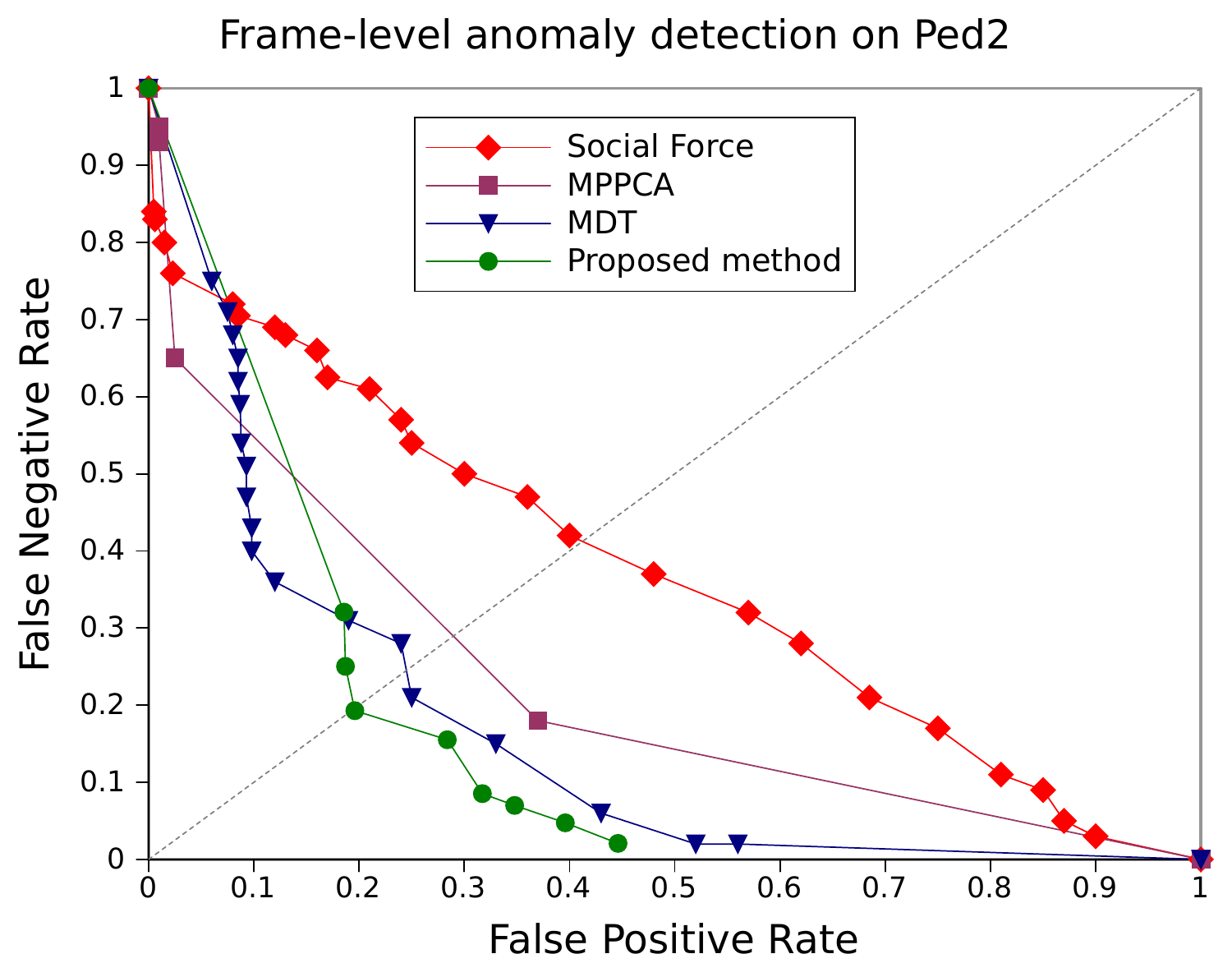}}
      \end{minipage}
    \end{minipage}
  
    \begin{minipage}{\textwidth}
      \centering
      \begin{minipage}{0.05\textwidth}
        ~
      \end{minipage}
      \hfill
      \begin{minipage}{0.90\textwidth}
      ~
      
      ~
      \hrule
      ~
      
      ~
      \end{minipage}
    \end{minipage}
  
    \begin{minipage}{\textwidth}
      \centering
      \begin{minipage}{0.05\textwidth}
        \centerline{\small\bf (c)}
      \end{minipage}
      \hfill
      \begin{minipage}{0.90\textwidth}
        \centerline{\includegraphics[width=1\textwidth]{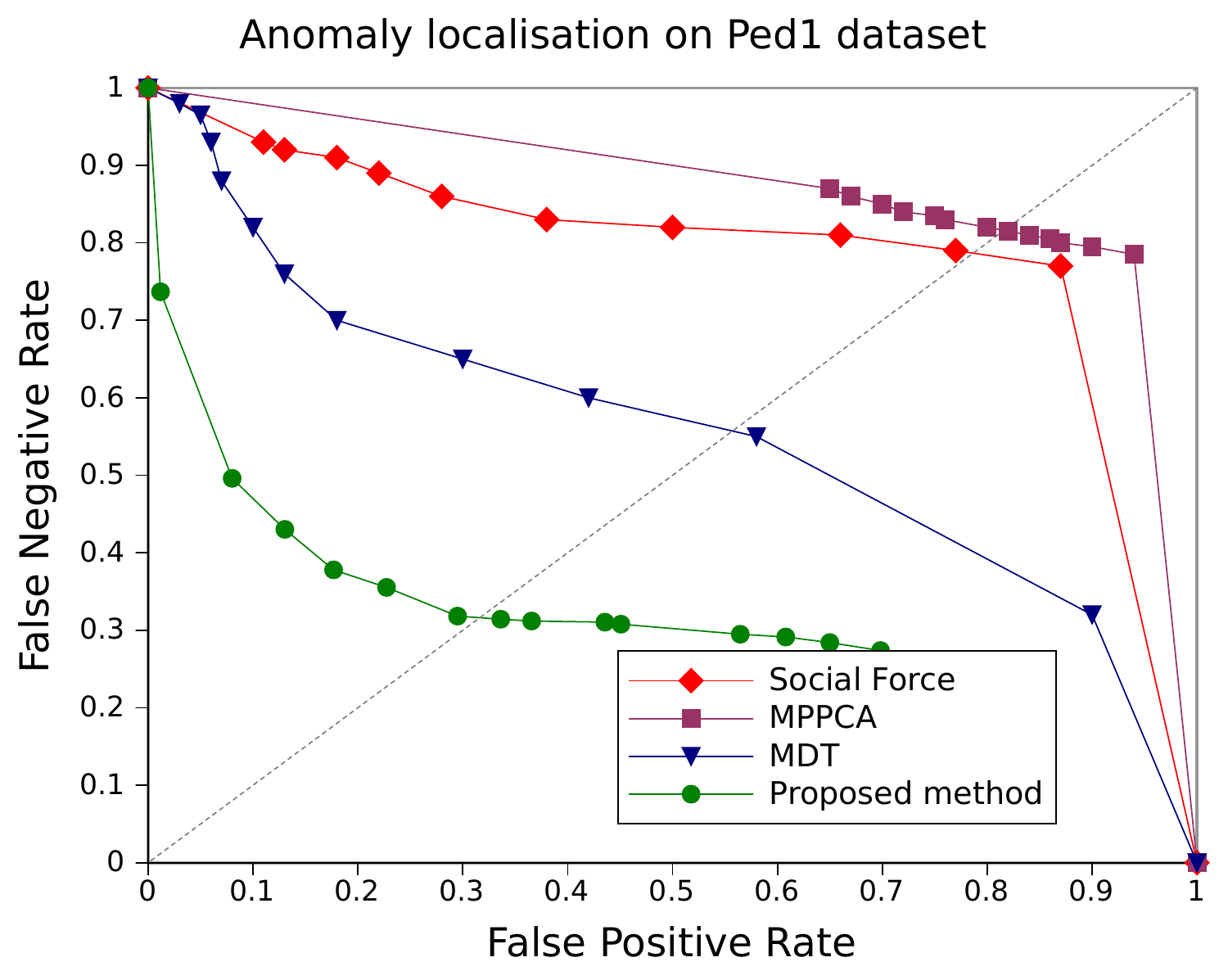}}
      \end{minipage}
    \end{minipage}

  \end{minipage}
  
  \vspace{1ex}
  
  \caption
    {
    ROC curves obtained on the UCSD Anomaly Detection dataset,
    with the bottom-left corner representing ideal performance:
    {\bf (a)}~frame-level anomaly detection on the Ped1~\mbox{subset};
    {\bf (b)}~frame-level anomaly detection on~Ped2;
    {\bf (c)}~within-frame anomaly localisation on Ped1.
    In all cases, the proposed method outperforms the other approaches at the equal error rate (EER) level,
    most notably on the anomaly localisation experiment.
    }
  \label{fig:ped1 roc}
\end{figure}

The quantitative results for the frame-level evaluation are shown
in Table~\ref{tab:frame_detection_results} and in Fig.~\ref{fig:ped1 roc}(a)-(b).
The results for the within-frame evaluation are shown
in Table~\ref{tab:frame_localisation_results} and in Fig.~\ref{fig:ped1 roc}(c).
Some of the qualitative results  obtained by the proposed method are shown in Fig.~\ref{fig:ad_sample_frames}. 
In Tables~\ref{tab:frame_detection_results} and~\ref{tab:frame_localisation_results},
the equal error rate (EER) is the point where the false negative rate is equal to the false positive rate.
At the EER,
the proposed method outperforms the other methods at both the frame-level and within-frame evaluations,
most notably on the anomaly localisation task.

An experimental implementation of the proposed algorithm in C++ yielded {\small $12$} fps ({\small $720$} frames per minute)
on a standard 3~GHz PC,
for sequences of images with a size of {\small $240 \times 160$} (ie.~processing the Ped1 subset).
We note that this is several orders of magnitude faster than the MDT method,
which takes 25 seconds to process each frame (ie.~{\small $2.4$} frames per minute)~\cite{mahadevan2010anomaly}.

\begin{table}[!tb]
  \centering
  \begin{small}
    \begin{tabular}{|c|c|c|c|c|} \hline
      \multirow{2}{*}{\bf Approach}    & Social Force          & MPPCA                     & MDT                 & Proposed  \\ 
                                     & \cite{mehran2009abnormal}   & \cite{10.1109/CVPRW.2009.5206569}   &\cite{mahadevan2010anomaly}        & method    \\ \hline \hline
      {\bf Ped1}             &  31\%       &      40\%                 &  25\%                         &\textbf{22.5}\%    \\ \hline
      {\bf Ped2}           &  42\%         &      30\%                 &  25\%                     &\textbf{20.0}\%  \\ \hline
      {\bf Average}            &  37\%         &      35\%                   &  25\%                         &\textbf{21.25}\%  \\ \hline
    \end{tabular}
  \end{small}
  \vspace{1ex}
  \caption
    {
    Equal error rates (EERs) for {\it frame-level} anomaly detection, obtained on the Ped1 and Ped2 subsets on the UCSD dataset.
    }
  \label{tab:frame_detection_results}
  
  \vspace{2.5ex}
  
  \begin{small}
    \begin{tabular}{|c|c|c|c|c|} \hline
      \multirow{2}{*}{\bf Approach}    & Social Force          & MPPCA                  & MDT                & Proposed  \\ 
                                       & \cite{mehran2009abnormal}   & \cite{10.1109/CVPRW.2009.5206569}  &\cite{mahadevan2010anomaly}     & method    \\ \hline \hline
      {\bf Ped1}           &      79\%             &  82\%                    &55\%                      &\textbf{32.0}\%       \\ \hline
    \end{tabular}
  \end{small}
  \vspace{1ex}
  \caption{ \small
     EERs for {\it pixel-level} anomaly localisation.
     }
  \label{tab:frame_localisation_results}
  \vspace{-1ex}
\end{table}


%

The proposed method has the ability to pick up anomalies (eg.~skateboarder, bike) 
present even at the far end of the scene (eg.~2nd and 3rd images in column~{\bf (a)}).
%
%
However, the last image in column {\bf (a)} contains a `miss':
the biker was not detected.
The cyclist was riding slowly and matching the pace of the neighbouring pedestrian (\mbox{bottom-left} \mbox{corner}). 
The texture in this context has strong vertical gradients making the biker appear as a pedestrian.
Using a more detailed texture descriptor may help in such cases.

We also note a false positive (a~pedestrian being detected as anomaly) in the last image of column {\bf (b)}.
Upon further investigation,
this false positive was due to the fact that the cells in the region of the pedestrian had minimal or no activity during the training phase.
Consequently any foreground object entering this zone during testing
was considered as anomalous,
irrespective of the observed features.

\renewcommand{\baselinestretch}{0.985}\small\normalsize
\section{Main Findings and Future Directions}
\label{sec:conclusion}
\vspace{-0.5ex}

In this paper we have proposed an anomaly detection algorithm targeted towards crowded scenes.
In addition to detecting anomalies based on motion,
it inspects for anomalies occurring due to size and texture.
Video images are initially segmented into foreground regions to confine the analysis to regions of interest,
ignoring the background (which might be dynamic).
Unlike most methods which compute the optical flow for all pixels or a fixed set of pixels, 
the flow is computed only for the foreground pixels,
thereby achieving a more precise estimate of object motion.

Features based on motion, size and texture are extracted at cell level (small fixed-size regions)
and are modelled independently for precise anomaly detection.
Motion and size features are modelled by an approximated kernel density estimation technique,
which is computationally efficient even on large training datasets.
The texture features are represented by an adaptively grown codebook,
which is generated in an online fashion.


Experiments on the recently published UCSD~Anomaly dataset 
(containing annotated~surveillance videos)
show that the proposed method obtains better results 
than several recent methods:
MPPCA, social force, mixture of dynamic textures (MDT).
The proposed method attained considerably more accurate anomaly localisation 
than the next best performing method, MDT,
while at the same time being several orders of magnitude faster than MDT.

As part of future work,
we aim to investigate the use of more descriptive features
such as the orientation of motion~\cite{10.1109/CVPRW.2009.5206569},
which would allow the detection of events such as wrong-way traffic.
It would also be useful to adaptively update the models over long periods of time,
allowing for context changes
(eg.~dense traffic might be usual during the day,
but it can be unusual at night).

The effect of the cell size should be analysed in the presence of object variations
due to factors such as image resolution,
perspective changes, as well as view angle.
The optimal cell size might be scene dependant and vary across the scene.
For example, in Fig~\ref{fig:ad_sample_frames}(a),
a larger size might be more appropriate for the bottom-left corner (where objects appear relatively large),
while a smaller size might be more effective in the top-right corner (where objects appear relatively small).

\renewcommand{\baselinestretch}{0.99125}\small\normalsize
\clearpage
\section*{Acknowledgements}

\begin{small}

NICTA is funded by the Australian Government
as represented by the {\it Department of Broadband, Communications and the Digital Economy},
as well as the Australian Research Council through the {\it ICT Centre of Excellence} program.
We thank Dr Mehrtash Harandi for useful discussions.

\end{small}

\balance
\begin{small}
\bibliographystyle{ieee}
\bibliography{references}

\begin{thebibliography}{10}\itemsep=-1pt

\bibitem{adam2008robust}
A.~Adam, E.~Rivlin, I.~Shimshoni, and D.~Reinitz.
\newblock {Robust real-time unusual event detection using multiple
  fixed-location monitors}.
\newblock {\em IEEE Trans. Pattern Analysis and Machine Intelligence},
  30(3):555--560, 2008.

\bibitem{bahlmann2005system}
C.~Bahlmann, Y.~Zhu, V.~Ramesh, M.~Pellkofer, and T.~Koehler.
\newblock {A system for traffic sign detection, tracking, and recognition using
  color, shape, and motion information}.
\newblock In {\em IEEE Intelligent Vehicles Symposium}, pages 255--260, 2005.

\bibitem{Baltieri_2010}
D.~Baltieri, R.~Vezzani, and R.~Cucchiara.
\newblock Fast background initialization with recursive {H}adamard transform.
\newblock In {\em Advanced Video and Signal Based Surveillance (AVSS)}, pages
  165--171, 2010.

\bibitem{basharat2008learning}
A.~Basharat, A.~Gritai, and M.~Shah.
\newblock {Learning object motion patterns for anomaly detection and improved
  object detection}.
\newblock In {\em Computer Vision and Pattern Recognition (CVPR)}, pages 1--8,
  2008.

\bibitem{bengio_icml_2005}
S.~Bengio, J.~Mari\'{e}thoz, and M.~Keller.
\newblock The expected performance curve.
\newblock In {\em Int. Conf Machine Learning (ICML), \mbox{Workshop} on ROC
  Analysis in Machine Learning}, 2005.

\bibitem{bishop2006pattern}
C.~Bishop.
\newblock {\em {Pattern Recognition and Machine Learning}}.
\newblock Springer, 2006.

\bibitem{bouguet1999pyramidal}
J.~Bouguet.
\newblock {Pyramidal Implementation of the Lucas Kanade Feature Tracker:
  Description of the algorithm}.
\newblock {\em Microprocessor Research Labs, Intel Corporation}, 1999.

\bibitem{Chan_MDT_2008}
A.~B. Chan and N.~Vasconcelos.
\newblock {Modeling, Clustering, and Segmenting Video with Mixtures of Dynamic
  Textures}.
\newblock {\em IEEE. Trans. Pattern Analysis and Machine Intelligence},
  30(5):909--926, 2008.

\bibitem{Chan_1979}
T.~F. Chan, G.~H. Golub, and R.~J. LeVeque.
\newblock Updating formulae and a pairwise algorithm for computing sample
  variances.
\newblock Technical Report STAN-CS-79-773, Department of Computer Science,
  Stanford University, 1979.

\bibitem{chandola2009anomaly}
V.~Chandola, A.~Banerjee, and V.~Kumar.
\newblock {Anomaly detection: A survey}.
\newblock {\em ACM Computing Surveys}, 41(3), 2009.

\bibitem{ermis2008motion}
E.~Ermis, V.~Saligrama, P.~Jodoin, and J.~Konrad.
\newblock {Motion segmentation and abnormal behavior detection via behavior
  clustering}.
\newblock In {\em Int. Conf. Image Processing (ICIP)}, pages 769--772, 2008.

\bibitem{Gonzalez_2008}
R.~Gonzalez and R.~Woods.
\newblock {\em {Digital Image Processing}}.
\newblock Pearson Prentice Hall, 3rd edition, 2008.

\bibitem{haering2008evolution}
N.~Haering, P.~Venetianer, and A.~Lipton.
\newblock {The evolution of video surveillance: an overview}.
\newblock {\em Machine Vision and Applications}, 19(5):279--290, 2008.

\bibitem{hu2006system}
W.~Hu, X.~Xiao, Z.~Fu, D.~Xie, T.~Tan, and S.~Maybank.
\newblock {A system for learning statistical motion patterns}.
\newblock {\em IEEE Trans. Pattern Analysis and Machine Intelligence},
  28(9):1450--1464, 2006.

\bibitem{10.1109/CVPRW.2009.5206569}
J.~Kim and K.~Grauman.
\newblock Observe locally, infer globally: A~space-time {MRF} for detecting
  abnormal activities with incremental updates.
\newblock In {\em Computer Vision and Pattern Recognition (CVPR)}, pages
  2921--2928, 2009.

\bibitem{kratz2009anomaly}
L.~Kratz and K.~Nishino.
\newblock {Anomaly detection in extremely crowded scenes using spatio-temporal
  motion pattern models}.
\newblock In {\em Computer Vision and Pattern Recognition (CVPR)}, pages
  1446--1453, 2009.

\bibitem{lee2002image}
T.~Lee.
\newblock {Image representation using 2D Gabor wavelets}.
\newblock {\em IEEE Trans. Pattern Analysis and Machine Intelligence},
  18(10):959--971, 1996.

\bibitem{lucas1981iterative}
B.~Lucas and T.~Kanade.
\newblock {An iterative image registration technique with an application to
  stereo vision}.
\newblock In {\em Int. Joint Conf. Artificial Intelligence}, volume~3, pages
  674--679, 1981.

\bibitem{mahadevan2010anomaly}
V.~Mahadevan, W.~Li, V.~Bhalodia, and N.~Vasconcelos.
\newblock {Anomaly detection in crowded scenes}.
\newblock In {\em Computer Vision and Pattern Recognition (CVPR)}, pages
  1975--1981, 2010.

\bibitem{mehran2009abnormal}
R.~Mehran, A.~Oyama, and M.~Shah.
\newblock {Abnormal crowd behavior detection using social force model}.
\newblock In {\em Computer Vision and Pattern Recognition (CVPR)}, pages
  935--942, 2009.

\bibitem{piciarelli2008trajectory}
C.~Piciarelli, C.~Micheloni, and G.~Foresti.
\newblock {Trajectory-based anomalous event detection}.
\newblock {\em IEEE Trans. Circuits and Systems for Video Technology},
  18(11):1544--1554, 2008.

\bibitem{Reddy_IVP_2011}
V.~Reddy, C.~Sanderson, and B.~C. Lovell.
\newblock A low complexity algorithm for static background estimation from
  cluttered image sequences in surveillance contexts.
\newblock {\em EURASIP Journal on Image and Video Processing}, 2011.
\newblock {\footnotesize
  \href{http://dx.doi.org/10.1155/2011/164956}{DOI:~10.1155/2011/164956}}.

\bibitem{reddy2010adaptive}
V.~Reddy, C.~Sanderson, A.~Sanin, and B.~C. Lovell.
\newblock Adaptive patch-based background modelling for improved foreground
  object segmentation and tracking.
\newblock In {\em Advanced Video and Signal Based Surveillance (AVSS)}, pages
  172--179, 2010.
\newblock {\footnotesize
  \href{http://dx.doi.org/10.1109/AVSS.2010.84}{DOI:~10.1109/AVSS.2010.84}}.

\bibitem{remagnino2001classifying}
P.~Remagnino and G.~Jones.
\newblock {Classifying Surveillance Events from Attributes and Behaviour}.
\newblock In {\em British Machine Vision Conference (BMVC)}, 2001.

\bibitem{saleemi2008probabilistic}
I.~Saleemi, K.~Shafique, and M.~Shah.
\newblock {Probabilistic modeling of scene dynamics for applications in visual
  surveillance}.
\newblock {\em IEEE Trans. Pattern Analysis and Machine Intelligence}, pages
  1472--1485, 2008.

\bibitem{saligrama2010video}
V.~Saligrama, J.~Konrad, and P.~Jodoin.
\newblock {Video Anomaly Identification}.
\newblock {\em IEEE Signal Processing Magazine}, 27(5):18--33, 2010.

\bibitem{Armadillo_2010}
C.~Sanderson.
\newblock Armadillo: An open source {C++} linear algebra library for fast
  prototyping and computationally intensive experiments.
\newblock Technical report, NICTA, 2010.
\newblock \href{http://arma.sourceforge.net}{http://arma.sourceforge.net}.

\bibitem{shotton2006textonboost}
J.~Shotton, J.~Winn, C.~Rother, and A.~Criminisi.
\newblock {Textonboost: Joint appearance, shape and context modeling for
  multi-class object recognition and segmentation}.
\newblock In {\em European Conference on Computer Vision (ECCV), Lecture Notes
  in Computer Science (LNCS)}, volume 3953, pages 1--15. Springer, 2006.

\bibitem{theodoridispattern}
S.~Theodoridis and K.~Koutroumbas.
\newblock {\em {Pattern Recognition}}.
\newblock Academic Press, 3rd edition, 2006.

\bibitem{trucco1998introductory}
E.~Trucco and A.~Verri.
\newblock {\em {Introductory Techniques for 3D Computer Vision}}.
\newblock Prentice Hall, 1998.

\bibitem{wang2006learning}
X.~Wang, K.~Tieu, and E.~Grimson.
\newblock {Learning semantic scene models by trajectory analysis}.
\newblock In {\em European Conference on Computer Vision (ECCV), Lecture Notes
  in Computer Science (LNCS)}, volume 3953, pages 110--123, 2006.

\end{thebibliography}
\end{small}

\end{document}